# Training Gaussian Mixture Models at Scale via Coresets

**Mario Lucic**                                                    lucic@inf.ethz.ch
*Department of Computer Science*
*ETH Zurich*
*Universitätstrasse 6, 8092 Zürich, Switzerland*

**Matthew Faulkner**                                              mnfaulk@gmail.com
*Department of Electrical Engineering and Computer Sciences*
*Caltech*
*1200 E California Blvd, Pasadena, California 91125*

**Andreas Krause**                                                krausea@ethz.ch
*Department of Computer Science*
*ETH Zurich*
*Universitätstrasse 6, 8092 Zürich, Switzerland*

**Dan Feldman**                                                   dannyf.post@gmail.com
*Department of Computer Science*
*University of Haifa*
*199 Aba Khoushy Ave. Mount Carmel, Haifa, Israel*

## Abstract

How can we train a statistical mixture model on a massive data set? In this work we show how to construct *coresets* for mixtures of Gaussians. A coreset is a weighted subset of the data, which guarantees that models fitting the coreset also provide a good fit for the original data set. We show that, perhaps surprisingly, Gaussian mixtures admit coresets of size polynomial in dimension and the number of mixture components, while being *independent* of the data set size. Hence, one can harness computationally intensive algorithms to compute a good approximation on a significantly smaller data set. More importantly, such coresets can be efficiently constructed both in distributed and streaming settings and do not impose restrictions on the data generating process. Our results rely on a novel reduction of statistical estimation to problems in computational geometry and new combinatorial complexity results for mixtures of Gaussians. Empirical evaluation on several real-world datasets suggests that our coreset-based approach enables significant reduction in training-time with negligible approximation error.

## 1. Introduction

We consider the problem of training statistical mixture models, in particular mixtures of Gaussians on massive data sets. In contrast to parameter estimation for models with compact sufficient statistics, mixture models generally require inference over latent variables, which in turn depends on the full data set. Such data sets are often distributed across a cluster of machines, or arrive in a data stream, and have to be processed with limited memory. In this paper, we show that Gaussian mixture models (GMMs) admit small *coresets*:

A weighted subset of the data which guarantees that models fitting the coreset will also provide a good fit for the original data set. As a result, solving the estimation problem on the coreset $\mathcal{C}$ is almost as good as solving the estimation problem on the large data set $\mathcal{X}$. Critically, we show that the size of these coresets is *independent* of the size of the data set.

We focus on training mixtures of $\lambda$-semi-spherical Gaussians, where the covariance matrix $\Sigma_i$ of each component $i \in [k]$ has eigenvalues bounded in $[\lambda, 1/\lambda]$. More formally, given a data set $\mathcal{X}$ of $n$ points in $\mathbb{R}^d$, some $\varepsilon > 0$ and an integer $k \geq 1$, one can efficiently construct a weighted set $\mathcal{C} \subseteq \mathcal{X}$ of $\Theta\big(d^4 k^6 \varepsilon^{-2} \lambda^{-4}\big)$ points, such that for any mixture of $k$ $\lambda$-semi-spherical Gaussians $\theta = [(w_i, \mu_i, \Sigma_i)]_{i=1}^k$ it holds that the log-likelihood $\ln P(\mathcal{X} \mid \theta)$ of $\mathcal{X}$ under $\theta$ is approximated by the (properly weighted) log-likelihood $\ln P(\mathcal{C} \mid \theta)$ of $\mathcal{C}$ under $\theta$ to arbitrary accuracy as $\varepsilon \to 0$. Moreover, these coresets can be efficiently constructed in parallel (using a merge-reduce style computation), as well as in the streaming setting using space and update time per point polynomial in $d$, $k$, $\lambda^{-1}$, $\varepsilon^{-1}$, $\log n$ and $\log(1/\delta)$.

## 2. Background and Problem Statement

We first discuss the problem of parameter estimation of Gaussian mixture models by maximum likelihood estimation. We then turn our attention to the problem of approximating the log-likelihood using a weighted subset of the data set and formally define the desired coreset property.

### 2.1 Fitting Gaussian Mixture Models by Maximum Likelihood Estimation

Given a data set $\mathcal{X} = \{x_1, \ldots, x_n\} \subset \mathbb{R}^d$ we consider fitting a mixture of Gaussians $\theta = [(w_1, \mu_1, \Sigma_1), \ldots, (w_k, \mu_k, \Sigma_k)]$, that is, the distribution

$$P(x \mid \theta) = \sum_{i=1}^k w_i \mathcal{N}(x; \mu_i, \Sigma_i)$$

where $w_1, \ldots, w_k \geq 0$ are the mixture weights and $\sum_i w_i = 1$. The $i$-th mixture component is modeled as a multivariate Normal distribution parametrized by mean $\mu_i \in \mathbb{R}^d$ and covariance $\Sigma_i \in \mathbb{R}^{d \times d}$,

$$\mathcal{N}(x; \mu_i, \Sigma_i) = \frac{1}{\sqrt{|2\pi\Sigma_i|}} \exp\left(-\frac{1}{2}(x - \mu_i)^T \Sigma_i^{-1}(x - \mu_i)\right).$$

Assuming the data was generated i.i.d., the negative log-likelihood of the data is

$$\mathcal{L}(\mathcal{X} \mid \theta) = -\sum_j \ln P(x_j \mid \theta),$$

and we wish to obtain the maximum likelihood estimate (MLE) of the parameters

$$\theta^* = \operatorname*{argmin}_{\theta \in \mathfrak{C}} \mathcal{L}(\mathcal{X} \mid \theta),$$

where $\mathfrak{C}$ is a set of constraints ensuring that degenerate solutions are avoided. Hereby, for a symmetric matrix $\mathbf{A}$, let $\operatorname{spec}(\mathbf{A})$ be the set of all eigenvalues of $\mathbf{A}$. We define $\mathfrak{C} = \mathfrak{C}_\lambda = \{\theta = [(w_i, \mu_i, \Sigma_i)]_{i=1}^k \mid \forall_i : \operatorname{spec}(\Sigma_i) \subseteq [\lambda, 1/\lambda]\}$ for some small $\lambda \in (0, 1)$.

## 2.2 Approximating the Log-likelihood

Ideally, we would like to obtain $(1 + \varepsilon)$-multiplicative approximation for the likelihood

$$\prod_{x \in \mathcal{X}} P(x \mid \theta)$$

which implies an additive $\varepsilon$ error for the sum of log-likelihoods. What kind of approximation accuracy may we hope to expect? Notice that there is a nontrivial issue of scale: Suppose we have a MLE $\theta^*$ for $\mathcal{X}$, and let $\alpha > 0$. Then straightforward linear algebra shows that we can obtain a MLE $\theta_\alpha^*$ for a scaled data set $\alpha D = \{\alpha x : x \in \mathcal{X}\}$ by simply scaling all means by $\alpha$, and covariance matrices by $\alpha^2$. For the log-likelihood, however, it holds that $\frac{1}{n}\mathcal{L}(\alpha D \mid \theta_\alpha^*) = d \ln \alpha + \frac{1}{n}\mathcal{L}(\mathcal{X} \mid \theta^*)$. Therefore, optimal solutions on one scale can be efficiently transformed to optimal solutions on a different scale, while maintaining the same *additive error*. Thus, we cannot expect to obtain a $(1 + \varepsilon)$-multiplicative approximation to the likelihood since any algorithm which achieves absolute error $\varepsilon$ at any scale could be used to compute parameter estimates (for means, covariances) with arbitrarily small error, simply by applying the algorithm to a scaled data set and transforming back the obtained solution.

An alternative, scale-invariant approach, may be to strive towards a *multiplicative error* $(1 + \varepsilon)$ for the sum of log-likelihoods. Unfortunately, this goal is also hard to achieve: Choosing a scaling parameter $\alpha$ such that $d \ln \alpha + \mathcal{L}(\mathcal{X} \mid \theta^*) = 0$ would require any algorithm that achieves any bounded multiplicative error to essentially incur *no error at all* when evaluating $\mathcal{L}(\alpha \mathcal{X} \mid \theta^*)$. The above observations hold even for the case $k = 1$ and $\Sigma = I$, where the mixture $\theta$ consists of a single Gaussian, and the log-likelihood is the sum of squared distances to a point $\mu$ and an additive term.

Motivated by the scaling issues discussed above, our goal is to approximate the data set $\mathcal{X}$ by a weighted set $C = \{(\gamma_1, \mathbf{x}_1'), \dots, (\gamma_m, \mathbf{x}_m')\} \subseteq \mathbb{R}_+ \times \mathbb{R}^d$ such that $\mathcal{L}(\mathcal{X} \mid \theta) \approx \mathcal{L}(C \mid \theta)$ for all $\theta \in \mathfrak{C}_\lambda$, where we define

$$\mathcal{L}(C \mid \theta) = -\sum_i \gamma_i \ln P(\mathbf{x}_i' \mid \theta).$$

The key idea is to decompose the negative log-likelihood into a data-dependent term, and a data-independent term, with the goal of approximating the latter with a coreset. To this end, we apply the following decomposition suggested by Arora and Kannan (2005):

$$\mathcal{L}(\mathcal{X} \mid \theta) = -\sum_{j=1}^n \ln \sum_{i=1}^k \frac{w_i}{\sqrt{|2\pi\Sigma_i|}} \exp\left(-\frac{1}{2}(x_j - \mu_i)^T \Sigma_i^{-1}(x_j - \mu_i)\right)$$

$$= -n \ln Z(\theta) + \text{cost}(\mathcal{X}, \theta),$$

where $Z(\theta) = \sum_i \frac{w_i}{\sqrt{|2\pi\Sigma_i|}}$ is a normalizer, and the function cost is defined as

$$\text{cost}(\mathcal{X}, \theta) = -\sum_{j=1}^n \ln \sum_{i=1}^k \frac{w_i}{Z(\theta)\sqrt{|2\pi\Sigma_i|}} \exp\left(-\frac{1}{2}(x_j - \mu_i)^T \Sigma_i^{-1}(x_j - \mu_i)\right).$$

Hereby, $Z(\theta)$ is a normalizing term which can be computed *exactly* and independently of the set $\mathcal{X}$. Furthermore, function $\text{cost}(\mathcal{X}, \theta)$ captures all dependencies of $\mathcal{L}(\mathcal{X} \mid \theta)$ on $\mathcal{X}$.

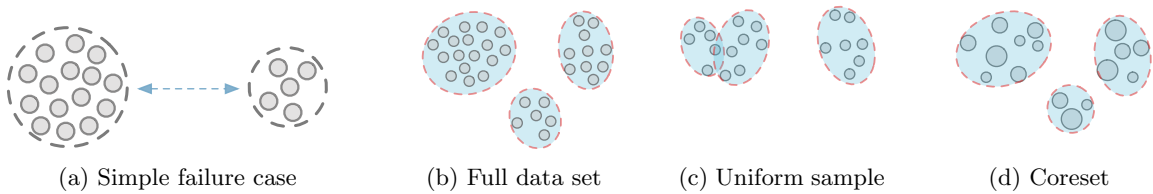

| (a) Simple failure case | (b) Full data set | (c) Uniform sample | (d) Coreset |

Figure 1: Figure (a) illustrates a simple example of well-separated Gaussians for which uniform subsampling fails arbitrarily badly. Consider two spherical Gaussian mixtures with weights $w_1 = 1/\sqrt{n}$ and $w_2 = 1 - 1/\sqrt{n}$. Unless the number of samples $m \in \Theta(\sqrt{n})$, the uniform sample will consist only of points from the first Gaussian, with high probability. Hence, moving the means of the Gaussians apart, the difference in MLEs on the full data set and the one on the uniform subsample can be made arbitrarily high. Figure (b) shows a GMM with 3 components fit on the full data set, (c) the model fit on the uniform subsample, and (d) the model fit on a coreset. The uniform subsample is likely to miss small clusters in presence of unbalanced data.

## 2.3 Coresets for Semi-spherical Gaussian Mixtures

We proceed by showing that it suffices to approximate $\text{cost}(\mathcal{X}, \theta)$ uniformly over $\theta \in \mathfrak{C}$.

**Definition 1** *We call a weighted data set $\mathcal{C}$ a $(k, \varepsilon)$-coreset for another (possibly weighted) set $\mathcal{X} \subset \mathbb{R}^d$, if for all mixtures $\theta \in \mathfrak{C}$ of $k$ Gaussians it holds that*

$$(1 - \varepsilon) \text{cost}(\mathcal{X}, \theta) \leq \text{cost}(\mathcal{C}, \theta) \leq (1 + \varepsilon) \text{cost}(\mathcal{X}, \theta).$$

Hereby, $\text{cost}(\mathcal{C}, \theta)$ is generalized to weighted data sets $\mathcal{C}$ in the natural way (weighing the contribution of each summand $\mathbf{x}'_j \in \mathcal{C}$ by its weight $\gamma_j$). Thus, as $\varepsilon \to 0$, for a sequence of $(k, \varepsilon)$-coresets $\mathcal{C}_\varepsilon$ we have that

$$\sup_{\theta \in \mathfrak{C}} |\mathcal{L}(\mathcal{C}_\varepsilon \mid \theta) - \mathcal{L}(\mathcal{X} \mid \theta)| = \sup_{\theta \in \mathfrak{C}} |\text{cost}(\mathcal{C}_\varepsilon, \theta) - \text{cost}(\mathcal{X} \mid \theta)| \to 0$$

which implies that $\mathcal{L}(\mathcal{C}_\varepsilon \mid \theta)$ uniformly approximates $\mathcal{L}(\mathcal{X} \mid \theta)$ (over $\theta \in \mathfrak{C}$).

The main motivation for constructing a $(k, \varepsilon)$-coreset $\mathcal{C}$ is to reduce the problem of fitting a mixture model on $\mathcal{X}$ to one of fitting a model on $\mathcal{C}$, since the optimal solution $\theta_\mathcal{C}$ is a good approximation (in terms of log-likelihood) of $\theta^*$. While finding the optimal $\theta_\mathcal{C}$ is a difficult problem, one can use a (weighted) variant of the EM algorithm to find a good solution. Moreover, if $|\mathcal{C}| \ll |\mathcal{X}|$, running EM on $\mathcal{C}$ is orders of magnitude faster than running EM on $\mathcal{X}$.

## 3. Efficient Coreset Construction

We start by contrasting the coreset approach with the "naive" approach of fitting the model on a uniform subsample. We show that the uniform subsampling approach can perform arbitrarily badly, while explicit worst-case guarantees can be given for the coreset based approach. We then present a simple coreset construction algorithm and conclude with a bound on the sufficient coreset size.

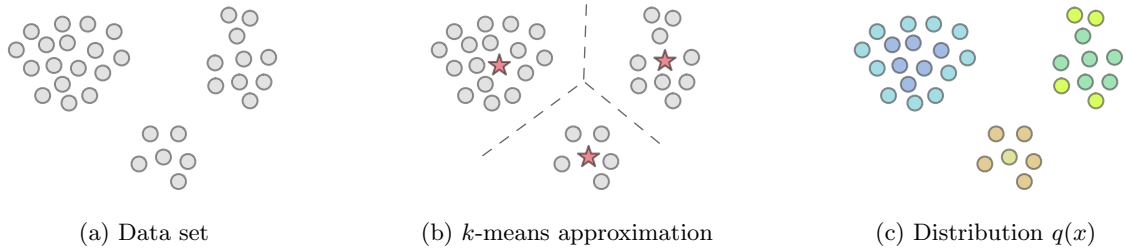

(a) Data set       (b) $k$-means approximation       (c) Distribution $q(x)$

Figure 2: Illustration of the coreset construction on a synthetic data set. Figure (a) shows the original data set and (b) the $k$-means bicriteria approximation. Figure (c) illustrates the computed sensitivity (blue-low, orange-high). The coreset sampling probabilities are inversely proportional to the size of the cluster which results in more representative samples.

### 3.1 Naive Approach via Uniform Sampling

A naive approach towards approximating the log-likelihood of $\mathcal{X}$ is to just pick a subset $\mathcal{C}$ uniformly at random. Unfortunately, such a simple strategy may result in arbitrary bad approximations of the log-likelihood as illustrated by the following example. Suppose the data set is generated from a mixture of two spherical Gaussians ($\Sigma_i = \mathbf{I}$) with weights $w_1 = \frac{1}{\sqrt{n}}$ and $w_2 = 1 - \frac{1}{\sqrt{n}}$. Unless $m = \Omega(\sqrt{n})$ points are sampled, with constant probability no data point generated from the second Gaussian is sampled. By moving the means of the Gaussians apart, $\mathcal{L}(\mathcal{X} \mid \theta_{\mathcal{C}})$ can be made arbitrarily worse than $\mathcal{L}(\mathcal{X} \mid \theta_{\mathcal{X}})$, where $\theta_{\mathcal{C}}$ and $\theta_{\mathcal{X}}$ are MLEs on $\mathcal{C}$ and $\mathcal{X}$ respectively. Thus, even for two well-separated Gaussians, uniform subsampling can perform arbitrarily poorly which is illustrated in Figure 2. The problem becomes even more pronounced as $k$ and $d$ grow.

This example already suggests that we must devise a sampling scheme that adaptively selects representative points from all "clusters" present in the data set. However, this implies that obtaining a coreset requires solving a chicken-and-egg problem, where we need to understand the density of the data to obtain the coreset, but simultaneously would like to use the coreset for density estimation.

### 3.2 Better Approximation via Importance Sampling

The key idea behind the coreset construction is that we can break the chicken-and-egg problem by first obtaining a rough approximation of the problem on $\mathcal{X}$ and use it to construct a non-uniform sampling scheme. This non-uniform sampling can be understood as an importance-weighted estimate of the log-likelihood $\mathcal{L}(\mathcal{X} \mid \theta)$, where the weights are optimized in order to reduce the variance. Feldman and Langberg (2011) successfully used the same idea to construct coresets for geometric clustering problems.

The critical insight is that, even though we seek to fit GMMs, it suffices to consider approximate solutions to the $k$-means clustering of $\mathcal{X}$ to construct a good sampling strategy. Intuitively, such a solution partitions $\mathcal{X}$ into Voronoi cells for which we can compute the density and the variance. Hence, one can bias the sampling towards less dense regions of $\mathcal{X}$. We prove that the resulting sampling strategy yields valid coresets in Appendix A.3. Our main result is the following Theorem which establishes the sufficient sample size which

---

**Algorithm 1** CORESET

1: **require:** Data set $\mathcal{X}$, bicriteria approximation $\mathcal{B}$, approximation factor $\alpha$, coreset size $m$.

2: **for** $j \leftarrow 1$ **to** $|\mathcal{B}|$

3:   $\mathcal{X}_j \leftarrow$ Set of points from $\mathcal{X}$ closest to point $\mathcal{B}_j$. Ties may be broken arbitrarily.

4: **for** $j \leftarrow 1$ **to** $|\mathcal{B}|$, **for each** $x \in \mathcal{X}_j$

5:   $s(x) \leftarrow \alpha \mathrm{d}(x, \mathcal{B})^2 + \frac{2\alpha}{|\mathcal{X}_j|} \sum_{x' \in \mathcal{X}_j} \mathrm{d}(x', \mathcal{B})^2 + \frac{2}{|\mathcal{X}_j|} \sum_{x' \in \mathcal{X}} \mathrm{d}(x', \mathcal{B})^2$

6: **for each** $x \in \mathcal{X}$

7:   $q(x) \leftarrow \frac{s(x)}{\sum_{x' \in \mathcal{X}} s(x')}$

8: $\mathcal{C} \leftarrow$ Sample $m$ weighted points from $\mathcal{X}$, where each point $x$ is sampled with probability

9:   $q(x)$ and assigned a weight $\frac{1}{m \cdot q(x)}$.

10: **return** $\mathcal{C}$

---

**Algorithm 2** K-MEANS++

1: **require:** Data set $\mathcal{X}$, number of clusters $k$.

2: $\mathcal{B} \leftarrow \{$Sample $x \in \mathcal{X}$ uniformly at random$\}$

3: **for** $j \leftarrow 2$ **to** $k$

4:   **for** $x$ **in** $\mathcal{X}$

5:     $p(x) \leftarrow \frac{\mathrm{d}^2(x, \mathcal{B})}{\sum_{x' \in \mathcal{X}} \mathrm{d}^2(x', \mathcal{B})}$

6:   Sample $x \in \mathcal{X}$ with probability $p(x)$ and add it to $\mathcal{B}$.

7: **return** $\mathcal{B}$

---

**Algorithm 3** ADAPTIVE SAMPLING

1: **require:** $\mathcal{X}$, $k$, failure probability $\delta$.

2: $R \leftarrow D, \mathcal{B} \leftarrow \emptyset, c \leftarrow \lceil 10dk \ln(1/\delta) \rceil$

3: **while** $|R| > c$

4:   $S \leftarrow$ Sample $c$ points uniformly from $R$

5:   $P \leftarrow \lceil |R|/2 \rceil$ points from $R$ closest to $S$

6:   $R \leftarrow R \setminus P$

7:   $\mathcal{B} \leftarrow \mathcal{B} \cup S$

8: $\mathcal{B} \leftarrow \mathcal{B} \cup R$

9: **return** $\mathcal{B}$

---

is independent of $n$, and only polynomial in $k$, $d$ and $\varepsilon$. For clarity, we define $\mathrm{d}(x, \mathcal{B}) = \min_{b \in \mathcal{B}} ||x - b||$ and $\phi(\mathcal{X}, \mathcal{B}) = \sum_{x \in \mathcal{X}} \mathrm{d}(x, \mathcal{B})^2$.

**Theorem 2** *Let* $\mathcal{X} \subset \mathbb{R}^d$, $\delta \in (0, 1)$, $\varepsilon \in (0, 1/2)$, $k \in \mathbb{N}$, *and* $\lambda \in (0, 1)$. *Let* $\mathcal{B}_1, \ldots, \mathcal{B}_p$ *be the outputs of* $p = \lceil \log_2 1/\delta \rceil$ *independent runs of Algorithm 2 with input* $\mathcal{X}$ *and* $k$. *Let* $\mathcal{C}$ *be the output of Algorithm 1 with* $\alpha = 16(\log_2 k + 2)$, $\mathcal{B}^\star = \mathrm{argmin}_{i \in \{1, \ldots, p\}} \phi(\mathcal{X}, \mathcal{B}_i)$ *and coreset size*

$$m \geq c \cdot \frac{d^4 k^6 + k^2 \log \frac{1}{\delta}}{\lambda^4 \varepsilon^2},$$

*where* $c > 0$ *is an absolute constant. Then, with probability at least* $1 - \delta$, *the set* $\mathcal{C}$ *is a* $(k, \varepsilon)$-*coreset of* $\mathcal{X}$.

To establish the resut we first reduce the MLE problem to the Euclidean space where the log-likelihood contribution of a data point given a model $\theta \in \mathfrak{C}$ can be expressed in a purely geometric manner. We then apply a coreset construction framework introduced by Feldman and Langberg (2011), which formalizes the intuition that one should sample the points with a potentially large impact more often. We then show that to compute the sampling distribution $q(x)$ it suffices to consider a rough approximation so the $k$-means clustering of $\mathcal{X}$. To guarantee uniform convergence over $\mathfrak{C}$ we bound the combinatorial complexity of the family of functions induced by the MLE of GMMs and show that it has a polynomial dependency on $k$ and $d$. The full proof is presented in the Appendix.

Several implementation choices are available. Firstly, we prove that one can use any $(\alpha, \beta)$-*bicriteria solution* to $k$-means ($\alpha$ approximate with respect to the optimal $k$-means clustering, using $\beta k$ centers). The suggested algorithm, `k-means++`, provides a solution $\mathcal{B}$ of size $k$ with approximation $\mathcal{O}(\log_2 k)$ in expectation in time $\mathcal{O}(nkd)$ and results in coresets of size *independent* of the data set size (Theorem 2). For other bicriteria approximation algorithms offering different tradeoffs in terms of computational complexity and the approximation guarantee we refer the reader to Bachem et al. (2016a,b), Makarychev et al. (2016) and Feldman and Langberg (2011) presented in Algorithm 3. Furthermore, all steps of the algorithm are parallelizable. The bicriteria approximation can be computed via $k$-means|| (Bahmani et al., 2012) and other quantities (i.e. $s(x)$, $q(x)$) can be computed in parallel. Finally, the importance sampling step can be implemented to run in constant-time per point with linear-time preprocessing (Vose, 1991).

## 4. Fitting a GMM on the Coreset using Weighted EM

Once the coreset $\mathcal{C}$ is constructed, we need to fit a mixture model that takes into account the point weights. Since the coreset size is independent of the cardinality of $\mathcal{X}$ we can (at least from the perspective of developing a polynomial time algorithm) afford to use a more "expensive" method. In geometric clustering problems such as $k$-means or $k$-median, where data points are hard-assigned to the closest cluster (point, subspace, etc.), it is possible to find the *optimal* clustering via exhaustive search, by simply considering all possible partitions of the coreset, and picking the best one. This procedure – constructing a coreset of size independent of $n$, and then using exhaustive search on the coreset – yields a (randomized) polynomial time approximation scheme (PTAS): It is guaranteed to achieve multiplicative error $1 + \varepsilon$, in time which is polynomial in $n$, but exponential in all other quantities (in particular $1/\varepsilon$). For mixture models, this exhaustive search algorithm is not feasible, since points are not hard-assigned to a cluster, but "soft-assigned" (according to the cluster membership probabilities). One approach, which we employ in our experiments, is to use a natural generalization of the EM algorithm, which takes the coreset weights into account. The details are presented in Algorithm 4 and the derivation of the EM update equations is presented in Appendix B.

Since the EM algorithm is applied on a significantly smaller data set, it can be initialized using multiple random restarts. In our experiments, we show that running weighted EM on the coreset typically leads to comparable performance (in terms of log-likelihood) as running EM on the full data set. A practical issue when using EM to fit a Gaussian mixture model is to ensure non-degeneracy which occurs when the MLE estimate for the (co)variance is zero implying infinite density and infinite log-likelihood (Bishop, 2006). The generally accepted remedy is to constrain the variances to be greater than some small apriori chosen value $\lambda > 0$ which is the strategy we employ in the experimental evaluation (Bishop, 2006). The alternative is to consider a fully Bayesian framework whereby one introduces a prior on the covariance matrices which induces large penalties for small values of the (co)variance in the resulting *maximum a posteriori* estimation problem (Murphy, 2012).

---

**Algorithm 4** EM for GMMs

---

1: **require:** Data set $\mathcal{X}$, point weights $\gamma$, number of components $k$, prior threshold $\lambda$.
2:   $\eta \leftarrow$ WEIGHTED-K-MEANS$(\mathcal{X}, \gamma, k)$    $\triangleright$ $\eta_{ij} = 1$ iff $x_i$ was assigned to cluster $j$.
3:   $w, \mu, \Sigma \leftarrow$ MAXIMIZATION$(\mathcal{X}, \eta, \lambda)$
4: **while not converged**
5:     $\eta \leftarrow$ EXPECTATION$(\mathcal{X}, \gamma, w, \mu, \Sigma)$
6:     $w, \mu, \Sigma \leftarrow$ MAXIMIZATION$(\mathcal{X}, \eta, \lambda)$
7: **return** $w, \mu, \Sigma$

---

**Algorithm 5** EXPECTATION

---

1: **require:** Data set $\mathcal{X}$, point weights $\gamma$, component weights $w$, means $\mu$, covariances $\Sigma$.
2:   $z \leftarrow 0_n$
3: **for** $i \in [n], j \in [k]$
4:     $\eta_{ij} \leftarrow w_j \mathcal{N}(x_i \mid \mu_j, \Sigma_j)$
5:     $z_i \leftarrow z_i + \eta_{ij}$
6: **for** $i \in [n], j \in [k]$
7:     $\eta_{ij} \leftarrow \gamma_i \eta_{ij} / z_i$
8: **return** $\eta$

---

**Algorithm 6** MAXIMIZATION

---

1: **require:** $\mathcal{X}$, responsibilities $\eta$, threshold $\lambda$.
2:   $z \leftarrow 0_k, \mu \leftarrow 0_{k \times d}, \Sigma \leftarrow 0_{k \times d \times d}$
3: **for** $j \in [k], i \in [n]$
4:     $z_j \leftarrow z_j + \eta_{ij}$
5:     $\mu_j \leftarrow \mu_j + \eta_{ij} x_j$
6:     $\Sigma_j \leftarrow \Sigma_j + \eta_{ij}(x_i - \mu_j)(x_i - \mu_j)^T$
7: **for** $j \in [k]$
8:     $w_j \leftarrow z_j / ||z||_1$
9:     $\mu_j \leftarrow \mu_j / z_j$
10:    $\Sigma_j \leftarrow \Sigma_j / z_j + I_\lambda$
11: **return** $(w, \mu, \Sigma)$

---

## 5. Streaming and Parallel Computation

One advantage of coresets is that they can be constructed in parallel, as well as in a streaming setting where data points arrive one by one, and it is impossible to remember the entire data set due to memory constraints. The key insight is that coresets satisfy certain composition properties, which have previously been used by Har-Peled and Mazumdar (2004) for streaming and parallel construction of coresets for geometric clustering problems such as $k$-median and $k$-means.

1. Let $\mathcal{C}_1$ be a $(k, \varepsilon)$-coreset for $\mathcal{X}_1$, and $\mathcal{C}_2$ be a $(k, \varepsilon)$-coreset for $\mathcal{X}_2$. Then $\mathcal{C}_1 \cup \mathcal{C}_2$ is a $(k, \varepsilon)$-coreset for $\mathcal{X}_1 \cup \mathcal{X}_2$.

2. Let $\mathcal{C}$ be a $(k, \varepsilon)$-coreset for $\mathcal{X}$, and $\mathcal{C}'$ be a $(k, \delta)$-coreset for $\mathcal{C}$. Then $\mathcal{C}'$ is a $(k, (1 + \varepsilon)(1 + \delta) - 1)$-coreset for $\mathcal{X}$.

### 5.1 Streaming Computation

In the streaming setting, we assume that points arrive one-by-one, but we do not have enough memory to remember the entire data set. Thus, we wish to maintain a coreset over time, while keeping only a small subset of $\mathcal{O}(\log n)$ coresets in memory, where $n$ is the number of points seen. The idea is to construct and store in memory a coreset for every block of $\text{poly}(d, k, \lambda^{-1}, \varepsilon^{-1})$ consecutive points arriving in a stream (Bentley and Saxe, 1980; Har-Peled and Mazumdar, 2004). When we have two $\varepsilon$-coresets in memory we first merge them which results in a $(k, \varepsilon)$-coreset via property (1). To maintain a small size, we compress them by computing a single coreset from the merged coresets via property (1) which increases the error. A naive approach that merges and compresses immediately as soon as two coresets have been constructed can incur an exponential increase in approximation error. Fortunately, it is possible to organize the merge-and-compress operations in a binary tree of height $\mathcal{O}(\log n)$, where we need to store in memory a single coreset for each level of the tree (Feldman et al., 2013b, Theorem 10.1.).

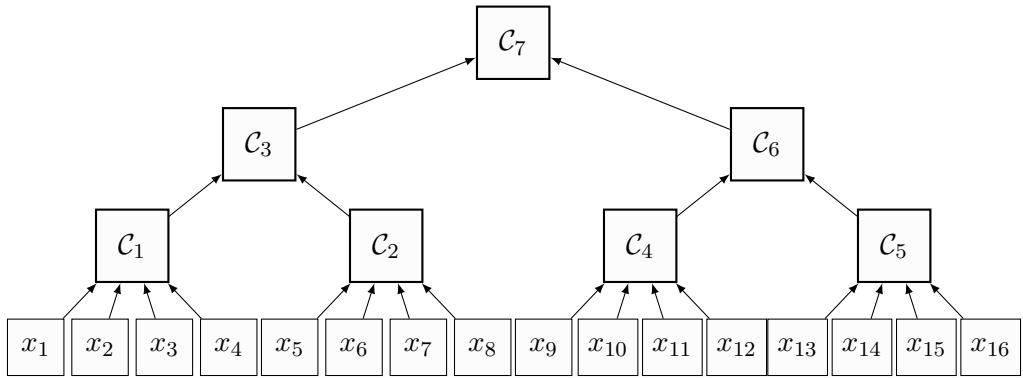

Figure 3: Coreset construction in the streaming setting. Black arrows indicate "merge-and-compress" operations. The (intermediate) coresets $\mathcal{C}_1, \dots, \mathcal{C}_7$ are enumerated in the order in which they would be generated in the streaming case. In the parallel case, $\mathcal{C}_1, \mathcal{C}_2, \mathcal{C}_4$ and $\mathcal{C}_5$ would be constructed in parallel, followed by parallel construction of $\mathcal{C}_3$ and $\mathcal{C}_6$, finally resulting in $\mathcal{C}_7$.

Consider Figure 3 which illustrates the tree computation for an example data set. In the following, $\varepsilon$-coreset denotes a $(\varepsilon, k)$-coreset ($k$ is fixed). In the first step, we construct a $\varepsilon$-coreset for $x_1, \dots, x_4$. We then construct a $\varepsilon$-coreset for $x_5, \dots, x_8$. At this point we have two $\varepsilon$-coresets and their union is, by property (1), a $\varepsilon$-coreset for $x_1, \dots, x_8$. By property (2) a coreset $\mathcal{C}_3$ of the union of those two coresets is a $4\varepsilon$-coreset for $x_1, \dots, x_8$ since $(1 + \varepsilon)^2 \leq 1 + 4\varepsilon$, for $0 \leq \varepsilon \leq 1$. Hence, we discard coresets $\mathcal{C}_1$ and $\mathcal{C}_2$ and keep only $\mathcal{C}_3$ in memory. We apply the same approach for $x_9, \dots, x_{16}$ and obtain $\mathcal{C}_6$, a $4\varepsilon$-coreset for $x_9, \dots, x_{16}$. Once again, we obtained two coresets at the same level of the tree ($\mathcal{C}_3$ and $\mathcal{C}_6$) and merging them we obtain a $4\varepsilon$-coreset for $x_1, \dots, x_{16}$. By property (2), the coreset of the union of $\mathcal{C}_3$ and $\mathcal{C}_6$ is a $13\varepsilon$-coreset for the whole data set since $(1 + 4\varepsilon)(1 + \varepsilon) \leq 1 + 13\varepsilon$. Finally, only coreset $C_7$ is kept in memory.

In general, to ensure an error of $\varepsilon$, it suffices that the intermediate coreset error is bounded by $\varepsilon' = \frac{\varepsilon}{6 \log_2 n}$ as the height of the tree is at most $\lceil \log_2 n \rceil$ and $(1 + \varepsilon')^{\lceil \log_2 n \rceil} \leq 1 + \frac{\varepsilon}{3}$. Thus, by property (2), a $(\varepsilon/3)$-coreset of the union of all coresets in memory has the approximation error bounded by $(1 + \varepsilon/3)^2 \leq 1 + \varepsilon$. As we do not know $n$ *a priori*, we compute coresets for data batches of exponentially increasing size – the total time and space requirements are dominated by the last batch whose size is upper bounded by $n$ (Feldman et al., 2013b).

**Theorem 3** *A $(k, \varepsilon)$-coreset for a stream of $n$ points in $\mathbb{R}^d$ can be computed for the $\lambda$-semi-spherical GMM using update time per point and memory $\mathrm{poly}(d, k, \lambda^{-1}, \varepsilon^{-1}, \log n, \log(1/\delta))$ with probability at least $1 - \delta$.*

In order to construct a coreset for the union of two (weighted) coresets, we use weighted versions of Algorithms 1 and 2, where we consider a weighted point as copies of a non-weighted point (possibly with fractional weight).

## 5.2 Distributed Computation

Using the same ideas from the streaming model, a (non-parallel) coreset construction can be transformed into a parallel one. We partition the data and compute a coreset for each partition independently. We then in parallel merge via property (1) two coresets, and compute a single coreset for every pair of such coresets exploiting the property (2). Continuing in this manner yields a process that takes $\mathcal{O}(\log n)$ iterations of parallel computation. This computation is naturally suited for map-reduce (Dean and Ghemawat, 2004) style computations, where the map tasks compute coresets for disjoint parts of $\mathcal{X}$, and the reduce tasks perform the merge-and-compress operations. Figure 3 illustrates this parallel construction.

**Theorem 4** *A $(k, \varepsilon)$-coreset for a set of $n$ points in $\mathbb{R}^d$ can be computed for the $\lambda$-semi-spherical GMM using $m$ machines in time $(n/m) \cdot \text{poly}(d, k, \lambda^{-1}, \varepsilon^{-1}, \log(1/\delta), \log n)$ with probability at least $1 - \delta$.*

Furthermore, if we have enough memory on one of the machines we can apply a simpler algorithm. First, partition the data to $m$ machines and compute a $(\varepsilon/3)$-coreset on each, producing coresets $\mathcal{C}_1, \mathcal{C}_2, \ldots, \mathcal{C}_m$. Then the union $C = \cup_{i=1}^m \mathcal{C}_i$ is a $(\varepsilon/3)$-coreset for the whole data set and its size is bounded by $m \cdot \max_i |\mathcal{C}_i|$. To obtain a coreset of size independent of the number of machines $m$, it suffices to construct a $(\varepsilon/3)$-coreset of $\mathcal{C}$.

Finally, Feldman and Tassa (2015) unified the streaming and distributed approaches when the data set size is unknown apriori. They propose splitting the input stream into several smaller streams whereby each machine proceeds to construct the coreset tree for its own stream.

## 6. Experimental Evaluation

The goal of this section is to demonstrate the effectiveness of using coresets for training Gaussian mixture models. To this end, we compare our coreset based approach to the "naive" approach of uniformly subsampling the data using models trained on the full data set as a baseline.

We construct coresets and uniform subsamples of sizes ranging from 1'000 to 10'000. For uniform subsampling, we subsample the data set and fit the model using Algorithm 4 where we set the weight of each point to one. For the coreset based approach with $k$ mixture components we first construct a bicriteria approximation using Algorithm 2 and construct the coreset using Algorithm 1. Finally, we fit a GMM on the weighed coreset using Algorithm 4. We stop iterating between EM steps if the number of iterations is greater than 100, or the relative log-likelihood changed is smaller than $10^{-3}$ and apply prior thresholding with $\lambda = 0.001$ (Section 4).

We compare the median results from 200 runs for the subsampling methods and *best* out of 100 runs on the full data set. For each run we store the sampling time, solving time and the log-likelihood $C$ on the test set. Finally, we calculate the *relative error*

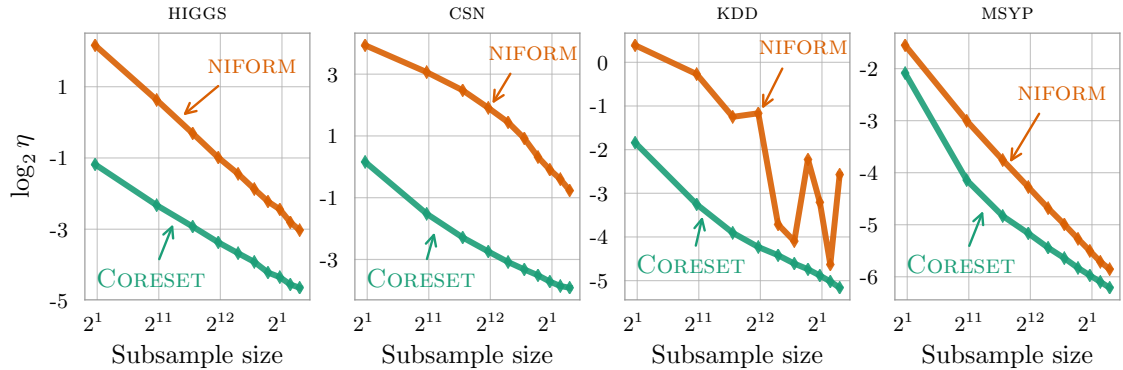

Figure 4: Median relative error $\eta$ with respect to the best models trained on the full data set (100 runs) with respect to the coreset/subsample (200 runs) For a fixed subsample size, models trained on coresets outperform models trained on the uniform subsample. Furthermore, we observe that the uniform subsampling may be much more sensitive to initialization (KDD).

$\eta = |(C - C_{full})/C_{full}|$ for both uniform subsampling and coresets. For each dataset we use 80% of the data for training and the remaining 20% for computing the error. We perform the evaluation on four real-world data sets and summarize our observations as follows:

1. HIGGS. Contains 11'000'000 instances describing signal processes which produce Higgs bosons and background processes which do not (Baldi et al., 2014). We consider the first two principal components and fit GMMs with 150 components. For $m = 10'000$ the coreset based approach leads to a speedup of $57.5\times$ with a relative error of 4%. At the same time, uniform subsampling leads to a relative error of 12.2%.

2. CSN. Contains 80'000 instances with 17 features extracted from acceleration data recorded from volunteers carrying and operating their phone in normal conditions (Faulkner et al., 2011). We fit GMM with 100 components. For $m = 10'000$ the coreset based approach leads to a speedup of $7\times$ with a relative error of 6.6%. At the same time, uniform subsampling leads to a relative error of 58.7%.

3. KDD. Contains 145'000 instances with 74 features measuring the match between a protein and a native sequence. We fit GMMs with 10 components. For $m = 10'000$ the coreset based approach leads to a speedup of $8.3\times$ with a relative error of 2.7%. At the same time, uniform subsampling leads to a relative error of 16.8%. We observe that models trained on the uniform subsample may be more sensitive to initialization.

4. MSYP. Contains 515'345 instances with 90 features which represent timbre average and timbre covariance for mostly western, commercial tracks ranging from 1922–2011. We consider the top 25 principal components and fit GMMs with 50 components. For $m = 5'000$ the coreset based approach leads to a speedup of $78\times$ with a relative error of 2.3%. At the same time, uniform subsampling leads to a relative error of 3.9%.

We observe that, for a fixed subsample size, coresets enjoy smaller approximation errors and obtain significant speedups with respect to solving the problem on the full data set.

## 7. Related Work

This paper is an extended version of Feldman et al. (2011) and provides significant improvements over the prior work. The theoretical analysis is executed directly on the negative log-likelihood function. This in turn leads to a novel importance sampling sampling strategy as well as a more practical algorithm with linear running time in $n$. Furthermore, we prove that one can use *any* bicriteria approximation for the $k$-means clustering problem as a basis for the importance sampling scheme. Overall, we are able to construct larger coresets in less time (approximately two orders of magnitude) and significantly improve the experimental results. To ensure the uniform convergence over $\mathfrak{C}$, we present new proof for the complexity (pseudo-dimension) of a mixture of Gaussians by forging a link to VC analysis of neural networks. The empirical evaluation considers data sets two orders of magnitude larger than those in prior work.

### 7.1 Learning Mixtures of Gaussians

The fundamental problem of learning Gaussian mixture models has received a great deal of interest. Dasgupta (1999) was the first to show that, given common covariance, bounded eccentricity, a bound on the smallest component weight, as well as a separation which scales with $\Omega(\sqrt{d})$, parameters of an unknown GMM $\theta$ can be estimated in polynomial time, with arbitrary accuracy $\varepsilon$, given i.i.d. samples from $\theta$. Subsequent works relax the assumption on separation to $d^{1/4}$ (Dasgupta and Schulman, 2000) and $k^{1/4}$ (Vempala and Wang, 2004). Feldman et al. (2006b) provide the first result that does not require any separation, but assumes that the Gaussians are axis-aligned. Moitra and Valiant (2010) and Belkin and Sinha (2010) prove that arbitrary GMMs with fixed number of components can be learned in polynomial time and sample complexity (but exponential dependence on $k$). Anandkumar et al. (2012, 2014) demonstrate that a spectral decomposition technique yields consistent parameter estimates from low-order observable moments, without additional separation assumptions. The aforementioned results hinge on non-degeneracy, that is, that the component means indeed span a $k$-dimensional subspace and the vector $w$ has strictly positive entries. The problem of fitting a mixture model with near-optimal log-likelihood for arbitrary data is studied by Arora and Kannan (2005). They provide a polynomial-time approximation scheme, provided that the Gaussians are identical spheres. In contrast, as detailed in Section 2, our results make only mild assumptions about the Gaussian components and allow one to explicitly trade-off the coreset size and the assumption strength. Critically, none of the algorithms described above applies to the streaming or parallel setting.

### 7.2 Approximation Algorithms via Coresets

Existence and construction of coresets have been investigated for a number of problems in computational geometry (Agarwal et al., 2005; Czumaj and Sohler, 2007) and have been used to great effect for a host of geometric and graph problems, including $k$-median (Har-Peled and Mazumdar, 2004), $k$-means (Feldman et al., 2007), $k$-center (Har-Peled and Varadarajan, 2004), $k$-line median (Feldman et al., 2006a), subspace approximation (Feldman et al., 2006a; Mahoney and Drineas, 2009), $(k, m)$-segment mean (Feldman et al., 2012), PCA and projective clustering (Feldman et al., 2013b), distributed $k$-means and $k$-median (Balcan et al., 2013), dictionary learning (Feldman et al., 2013a), $k$-segmentation of streaming data (Rosman et al., 2014), non-parametric estimation (Bachem et al., 2015), and clustering with Bregman divergences (Lucic et al., 2016b). A framework that generalizes and improves several of these results has appeared in Feldman and Langberg (2011). Notably, coresets also imply streaming algorithms for many of these problems (Har-Peled and Mazumdar, 2004; Agarwal et al., 2005; Frahling and Sohler, 2005; Feldman et al., 2007). Recently, coresets were leveraged to establish a space-time-data-risk tradeoff in the context of unsupervised learning (Lucic et al., 2015). Promising results in the context of empirical risk minimization have been demonstrated by Reddi et al. (2015). This paper uses the sensitivity framework of Feldman and Langberg (2011) and the quadratic dependency on the total sensitivity can be reduced to near-linear via Braverman et al. (2016). For a survey of the recent results we refer the reader to Bachem et al. (2017b); Phillips (2016).

## 8. Conclusion

We have shown how to efficiently construct coresets for estimating parameters of Gaussian mixture models by exploiting a connection between statistical estimation and clustering problems in computational geometry. We prove existence of coresets of size *independent* of the original data set size. To our knowledge, our results provide the first rigorous guarantees for obtaining compressed $\varepsilon$-approximations of the log-likelihood of mixture models for large data sets. The coreset construction algorithm is based on a simple importance sampling scheme and and has linear running time in $n$. We demonstrate that, by exploiting certain closure properties of coresets, it is possible to construct them in parallel, or in a single pass through a stream of data, using only $\mathrm{poly}(d, k, \lambda^{-1}, \varepsilon^{-1}, \log n, \log(1/\delta))$ space and update time. Critically, our coresets provide guarantees for any given (possibly unstructured) data, without assumptions on the distribution or model that generated it. In an empirical evaluation on several real-world datasets we observe a reduction in computational time of *up to two orders of magnitude*, while achieving a hold-out set likelihood competitive with the models trained on the full data set.

There are two interesting open problems: Is it possible to compute a coreset (i) for any set of $k$ mixture of Gaussians, i.e., whose size is independent of $\lambda$, and (ii) of size independent of $d$, as in Barger and Feldman (2016).

## Acknowledgments

We thank Olivier Bachem for invaluable discussions, suggestions and comments. This research was partially supported by ONR grant N00014-09-1-1044, NSF grants CNS-0932392, IIS-0953413, DARPA MSEE grant FA8650-11-1-7156, and the Zurich Information Security Center.

## Appendix A. Bounding the Coreset Size

Let $f : \mathcal{X} \times \mathfrak{C} \to \mathbb{R}_+$ be defined as

$$f_\theta(x) = -\ln \left( \sum_{i=1}^k \frac{w_i}{Z(\theta)\sqrt{|2\pi\Sigma_i|}} \exp\left(-\frac{1}{2}\left\|\Sigma_i^{-1/2}(x - \mu_i)\right\|_2^2\right) \right),$$

where $Z(\theta) = \sum_i w_i / \sqrt{|2\pi\Sigma_i|}$. For a point $x \in \mathcal{X}$ and a *query* $\theta$ – a parametrization of the mixture model – $f_\theta(x)$ measures the contribution of the point $x$ to the log-likelihood. Intuitively, to select the points for the coreset, we would like to perform importance sampling on points $x \in \mathcal{X}$, such that we sample points proportionally to the impact on the log-likelihood. Langberg and Schulman (2010) show that it suffices to perform importance sampling with respect to *sensitivity* – the worst-case contribution of a point.

**Definition 5 (Sensitivity)** *Sensitivity of $x \in \mathcal{X}$ w.r.t. $\mathcal{F} = \{f_\theta(\cdot) \mid \theta \in \mathfrak{C}\}$ is defined as*

$$\sigma_{\mathfrak{C}}(x) = \sup_{\theta \in \mathfrak{C}} \frac{f_\theta(x)}{\frac{1}{|\mathcal{X}|}\sum_{x' \in \mathcal{X}} f_\theta(x')}.$$

*Total sensitivity is defined as $\mathfrak{S} = \frac{1}{|\mathcal{X}|}\sum_{x \in \mathcal{X}} \sigma_{\mathfrak{C}}(x)$.*

For the Gaussian mixture model, the *query space* $\mathfrak{C}$ is the space of all possible GMMs with $k$ components in $d$ dimensions. While the exact sensitivity is hard to compute, Langberg and Schulman (2010) show that any uniform upper bound $s_{\mathfrak{C}}(x)$ to $\sigma_{\mathfrak{C}}(x)$ can be used. Looser bounds will lead to larger coresets, so one should aim to provide the tightest bound possible.

### A.1 Total Sensitivity by Reduction to the Euclidean Space

The following lemma bounds the difference in log-likelihood contribution of two points for any fixed $\lambda$-semi-spherical GMM.

**Lemma 6** *Let $\mathfrak{C}_\lambda$ be the family of $\lambda$-semi-spherical Gaussian mixtures of $k$ components. For a fixed $\theta \in \mathfrak{C}_\lambda$ define $f_\theta : \mathcal{X} \to [0, +\infty)$ as*

$$f_\theta(x) = -\ln \sum_{i=1}^k \frac{w_i}{Z(\theta)\sqrt{|2\pi\Sigma_i|}} \exp\left(-\frac{1}{2}(x - \mu_i)^T \Sigma_i^{-1}(x - \mu_i)\right),$$

*where $Z(\theta) = \sum_{i=1}^k \frac{w_i}{\sqrt{|2\pi\Sigma_i|}}$. Then, for every $x, y \in \mathbb{R}^d$ it holds that*

$$f_\theta(x) \le \frac{1}{\lambda}\|x - y\|_2^2 + 2f_\theta(y).$$

**Proof** Let $a, x \in \mathbb{R}^d$. By the weak triangle inequality for a fixed $i \in [k]$,

$$\left\|\Sigma_i^{-1/2}(x - \mu_i)\right\|_2^2 \le 2\left\|\Sigma_i^{-1/2}(x - a)\right\|_2^2 + 2\left\|\Sigma_i^{-1/2}(a - \mu_i)\right\|_2^2. \tag{1}$$

Let $UDU^T$ denote the SVD of $\Sigma_i$ and note that $U$ is an orthogonal matrix. As such,

$$\left\|\Sigma_i^{-1/2}(x-a)\right\|_2 = \left\|UD^{-1/2}U^T(x-a)\right\|_2 = \left\|D^{-1/2}U^T(x-a)\right\|_2$$
$$\leq \frac{\left\|U^T(x-a)\right\|_2}{\sqrt{\lambda}} \leq \frac{\|x-a\|_2}{\sqrt{\lambda}},$$

which combined with (1) yields

$$\left\|\Sigma_i^{-1/2}(x-\mu_i)\right\|_2^2 \leq \frac{2}{\lambda}\|x-a\|_2^2 + 2\left\|\Sigma_i^{-1/2}(a-\mu_i)\right\|_2^2.$$

As such,

$$f_\theta(x) \leq -\ln\left(\sum_{i=1}^k \frac{w_i}{Z(\theta)\sqrt{|2\pi\Sigma_i|}}\exp\left(-\frac{1}{\lambda}\|x-a\|_2^2 - \left\|\Sigma_i^{-1/2}(a-\mu_i)\right\|_2^2\right)\right)$$
$$= \frac{1}{\lambda}\|x-a\|_2^2 - \ln\left(\sum_{i=1}^k \frac{w_i}{Z(\theta)\sqrt{|2\pi\Sigma_i|}}\exp\left(-\frac{1}{2}\left\|\Sigma_i^{-1/2}(a-\mu_i)\right\|_2^2\right)^2\right)$$
$$\leq \frac{1}{\lambda}\|x-a\|_2^2 - \ln\left(\sum_{i=1}^k \frac{w_i}{Z(\theta)\sqrt{|2\pi\Sigma_i|}}\exp\left(-\frac{1}{2}\left\|\Sigma_i^{-1/2}(a-\mu_i)\right\|_2^2\right)\right)^2$$
$$= \frac{1}{\lambda}\|x-a\|_2^2 + 2f_\theta(a),$$

by Jensen's inequality and the fact that

$$\sum_{i=1}^k \frac{w_i}{Z(\theta)\sqrt{|2\pi\Sigma_i|}} = \frac{1}{Z(\theta)}\sum_{i=1}^k \frac{w_i}{\sqrt{|2\pi\Sigma_i|}} = 1.$$

∎

The critical insight necessary to compute an upper-bound on the sensitivity of each point is to note that the denominator in (5) can be lower-bounded by a rough approximation to the optimal $k$-means clustering of $\mathcal{X}$.

**Lemma 7** *Let $\mathfrak{C}_\lambda$ be the family of $\lambda$-semi-spherical Gaussian mixtures with $k$ components. Let $\mathcal{X} \subset \mathbb{R}^d$, and $C^\star = \min_{C \subset \mathbb{R}^{d \times k}} \phi(\mathcal{X}, C)$. Let $\mathcal{B} \subset \mathbb{R}^{d \times \beta}$ such that $\phi(\mathcal{X}, \mathcal{B}) \leq \alpha\phi(\mathcal{X}, C^\star)$. The sensitivity of $x \in \mathcal{X}$ with respect to $\mathfrak{C}_\lambda$,*

$$\sigma(x) = \sup_{\theta \in \mathfrak{C}_\lambda} \frac{f_\theta(x)}{\frac{1}{|\mathcal{X}|}\sum_{x' \in \mathcal{X}} f_\theta(x')},$$

*is bounded by*

$$\sigma(x) \leq s(x) = |\mathcal{X}|\frac{2}{\lambda^2}\left(\frac{\alpha\mathrm{d}(x,\mathcal{B})^2}{\sum_{x' \in \mathcal{X}}\mathrm{d}(x',\mathcal{B})^2} + \frac{2\alpha}{|\mathcal{X}_j|}\frac{\sum_{x' \in \mathcal{X}_j}\mathrm{d}(x',\mathcal{B})^2}{\sum_{x' \in \mathcal{X}}\mathrm{d}(x',\mathcal{B})^2} + \frac{2}{|\mathcal{X}_j|}\right).$$

*Furthermore, it holds that*

$$\mathfrak{S} \leq \frac{1}{|\mathcal{X}|}\sum_{x \in \mathcal{X}} s(x) = \frac{1}{\lambda^2}(6\alpha + 4\beta).$$

**Proof** Set $\mathcal{B}$ partitions $\mathcal{X}$ into $\beta$ Voronoi cells, $\mathcal{X}_1, \ldots, \mathcal{X}_\beta$. Consider some $x \in \mathcal{X}_j$ and $\theta \in \mathfrak{C}_\lambda$. By Lemma 6 it holds that

$$\frac{f_\theta(x)}{\sum_{x' \in \mathcal{X}} f_\theta(x')} \leq \underbrace{\frac{1}{\lambda} \frac{\mathrm{d}(x, \mathcal{B})^2}{\sum_{x' \in \mathcal{X}} f_\theta(x')}}_{\mathtt{u}} + \underbrace{\frac{2 f_\theta(y)}{\sum_{x' \in \mathcal{X}} f_\theta(x')}}_{\mathtt{v}} . \tag{2}$$

We now upper-bound the right hand side. To bound $\mathtt{u}$, let $\Sigma_i = UDU^T$ be the singular value decomposition of $\Sigma_i$ and $\mu_i$ the corresponding mean. Since $U$ is a rotation matrix

$$\left\| \Sigma_i^{-1/2}(x - \mu_i) \right\|_2 = \left\| UD^{-1/2}U^T(x - \mu_i) \right\|_2 = \left\| D^{-1/2}U^T(x - \mu_i) \right\|_2$$
$$\geq \sqrt{\lambda} \left\| U^T(x - \mu_i) \right\|_2 = \sqrt{\lambda} \left\| x - \mu_i \right\|_2 \geq \sqrt{\lambda} \mathrm{d}(x, \mu)$$

where $\mu = \{\mu_1, \ldots, \mu_k\}$. Noting that $Z(\theta)$ is a normalization constant it follows that

$$f_\theta(x) \geq -\ln \left( \sum_{i=1}^{k} \frac{w_i}{Z(\theta)\sqrt{|2\pi \Sigma_i|}} \exp\left( -\frac{\lambda}{2} \mathrm{d}(x, \mu)^2 \right) \right) = \frac{\lambda}{2} \mathrm{d}(x, \mu)^2$$

Summing over $x' \in \mathcal{X}$ yields

$$\sum_{x' \in \mathcal{X}} f_\theta(x') \geq \frac{\lambda}{2} \sum_{x' \in \mathcal{X}} \mathrm{d}(x, \mu)^2 \geq \frac{\lambda}{2} \min_{C \subset \mathbb{R}^{d \times k}} \sum_{x' \in \mathcal{X}} \mathrm{d}(x', C)^2 \geq \frac{\lambda}{2\alpha} \sum_{x' \in \mathcal{X}} \mathrm{d}(x', \mathcal{B})^2,$$

where the last inequality follows by the definition of $\mathcal{B}$. Thus,

$$\mathtt{u} = \frac{1}{\lambda} \frac{\mathrm{d}(x, \mathcal{B})^2}{\sum_{x' \in \mathcal{X}} f_\theta(x')} \leq \frac{2\alpha}{\lambda^2} \frac{\mathrm{d}(x, \mathcal{B})^2}{\sum_{x' \in \mathcal{X}} \mathrm{d}(x', \mathcal{B})^2}. \tag{3}$$

To upper-bound $\mathtt{v}$ consider again the Voronoi partitioning of $\mathcal{X}$ induced by $\mathcal{B}$. Let $x \in \mathcal{X}$ with the corresponding cell $\mathcal{X}_j \subseteq \mathcal{X}$, and $y = \mathcal{B}_j$ such that $\mathrm{d}(x, \mathcal{B}) = \|x - y\|_2$ ($y$ induced the cell $\mathcal{X}_j$). By swapping $x$ and $y$ in Lemma 6 we have

$$f_\theta(y) \leq \frac{1}{\lambda} \mathrm{d}(x, \mathcal{B})^2 + 2 f_\theta(x)$$

and summing over $x' \in \mathcal{X}_j$ yields

$$|\mathcal{X}_j| f_\theta(y) \leq \frac{1}{\lambda} \sum_{x' \in \mathcal{X}_j} \mathrm{d}(x', \mathcal{B})^2 + 2 \sum_{x' \in \mathcal{X}_j} f_\theta(x')$$
$$\leq \frac{1}{\lambda} \sum_{x' \in \mathcal{X}_j} \mathrm{d}(x', \mathcal{B})^2 + 2 \sum_{x' \in \mathcal{X}} f_\theta(x').$$

where the last inequality follows from $f_\theta(x) \geq 0$. Hence,

$$\frac{f_\theta(y)}{\sum_{x' \in \mathcal{X}} f_\theta(x')} \leq \frac{1}{\lambda} \frac{\sum_{x' \in \mathcal{X}_j} \mathrm{d}(x', \mathcal{B})^2}{|\mathcal{X}_j| \sum_{x' \in \mathcal{X}} f_\theta(x')} + \frac{2}{|\mathcal{X}_j|}$$
$$\leq \frac{2\alpha}{\lambda^2} \frac{\sum_{x' \in \mathcal{X}_j} \mathrm{d}(x', \mathcal{B})^2}{|\mathcal{X}_j| \sum_{x' \in \mathcal{X}} \mathrm{d}(x', \mathcal{B})^2} + \frac{2}{|\mathcal{X}_j|}. \tag{4}$$

Since the choice of $\theta$ was arbitrary, by applying the obtained bounds (3) and (4) to (2) it follows that

$$
\begin{aligned}
\sigma(x) &\leq |\mathcal{X}| \frac{2}{\lambda^2} \left( \frac{\alpha \mathrm{d}(x, \mathcal{B})^2}{\sum_{x' \in \mathcal{X}} \mathrm{d}(x', \mathcal{B})^2} + \frac{2\alpha}{|\mathcal{X}_j|} \frac{\sum_{x' \in \mathcal{X}_j} \mathrm{d}(x', \mathcal{B})^2}{\sum_{x' \in \mathcal{X}} \mathrm{d}(x', \mathcal{B})^2} + \frac{2\lambda^2}{|\mathcal{X}_j|} \right) \\
&\leq |\mathcal{X}| \frac{2}{\lambda^2} \left( \frac{\alpha \mathrm{d}(x, \mathcal{B})^2}{\sum_{x' \in \mathcal{X}} \mathrm{d}(x', \mathcal{B})^2} + \frac{2\alpha}{|\mathcal{X}_j|} \frac{\sum_{x' \in \mathcal{X}_j} \mathrm{d}(x', \mathcal{B})^2}{\sum_{x' \in \mathcal{X}} \mathrm{d}(x', \mathcal{B})^2} + \frac{2}{|\mathcal{X}_j|} \right) \\
&:= s(x)
\end{aligned}
$$

since $\lambda \in (0, 1)$. Hence, the total sensitivity may be bounded by

$$
\mathfrak{S} \leq \frac{1}{|\mathcal{X}|} \sum_{x \in \mathcal{X}} s(x) = \frac{1}{\lambda^2} (6\alpha + 4\beta).
$$

∎

An attractive property of this result is that the total sensitivity is independent of $|\mathcal{X}|$. Furthermore, to compute the bound we only need a bicriteria approximation to the $k$-means objective which can be computed in linear time with the popular K-MEANS++ algorithm (Arthur and Vassilvitskii, 2007) for which $\alpha = \mathcal{O}(\log_2 k)$ and $\beta = k$ resulting in total sensitivity of $\mathcal{O}(k/\lambda^2)$. As shown in Lucic et al. (2016a), this bound on the total sensitivity is tight (up to a constant) as there exists a data set $\mathcal{X}$ for which $\mathfrak{S} \in \Theta(k)$.

### A.2 Pseudo-dimension of Gaussian Mixtures

The other key factor in bounding the coreset size is the combinatorial complexity of the function family $\mathcal{F}$ induced by the maximum likelihood estimation of the Gaussian mixture. More complex models require more samples for uniform convergence. We first introduce the required definitions to ensure that the exposition is self-contained.

**Definition 8 (VC dimension)** *Let $\mathcal{X}$ be a ground set and $\mathcal{F}$ be a set of functions from $\mathcal{X}$ to $\{0, 1\}$. Fix a set $S = \{x_1, \ldots, x_n\} \subset \mathcal{X}$ and a function $f \in \mathcal{F}$. We call $S_f = \{x_i \in S \mid f(x_i) = 1\}$ the induced subset of $S$ by $f$. A subset $S = \{x_1, \ldots, x_n\}$ of $\mathcal{X}$ is shattered by $\mathcal{F}$ if $|\{S_f \mid f \in \mathcal{F}\}| = 2^n$. VC dimension of $\mathcal{F}$ is the size of the largest subset of $\mathcal{X}$ shattered by $\mathcal{F}$. If $\mathcal{F}$ can shatter sets of arbitrary size VC dimension of $\mathcal{F}$ is $\infty$.*

These notions naturally extend to functions mapping to $\mathbb{R}$ (or a subset thereof).

**Definition 9 (Pseudo-dimension)** *Let $\mathcal{X}$ be a ground set and $\mathcal{F}$ be a set of functions from $\mathcal{X}$ to the interval $[0, 1]$. Fix a set $S = \{x_1, \ldots, x_n\} \subset \mathcal{X}$, a set of reals $R = \{r_1, \ldots, r_n\}, r_i \in [0, 1]$ and a function $f \in \mathcal{F}$. We call $S_f = \{x_i \in S \mid f(x_i) \geq r_i\}$ the induced subset of $S$ formed by $f$ and $R$. Subset $S$ with associated values $R$ is shattered by $\mathcal{F}$ if $|\{S_f \mid f \in \mathcal{F}\}| = 2^n$. Pseudo-dimension of $\mathcal{F}$ is the cardinality of the largest shattered subset of $\mathcal{X}$. If $\mathcal{F}$ can shatter sets of arbitrary size pseudo-dimension of $\mathcal{F}$ is $\infty$.*

Clearly, for every space of a given pseudo-dimension we can construct a space with the same VC dimension as formalized by the following lemma.

**Lemma 10** *For any $f \in \mathcal{F}$ let $B_f$ be the indicator function of the region below or on the graph of $f$, i.e. $B_f(x, y) = \text{sgn}(f(x) - y)$. The pseudo-dimension of $\mathcal{F}$ is precisely the VC-dimension of the subgraph class $B_{\mathcal{F}} = \{B_f \mid f \in \mathcal{F}\}$.*

For the Gaussian mixture model, each $f_\theta$ consists exclusively of applications of the exponential function and arithmetic operations on real numbers. The problem of bounding the combinatorial complexity of such function classes has received a great deal of interest which culminated in the following result (Anthony and Bartlett, 2009, Theorem 8.14):

**Theorem 11** *Let $h$ be a function from $\mathbb{R}^m \times \mathbb{R}^d$ to $\{0, 1\}$, determining the class*

$$\mathcal{H} = \{x \mapsto h(x) : \theta \in \mathbb{R}^m\}.$$

*Suppose that $h$ can be computed by an algorithm that takes as input the pair $(\theta, x) \in \mathbb{R}^m \times \mathbb{R}^d$ and returns $h(\theta, x)$ after no more than $t$ of the following operations:*

(i) *the exponential function $\alpha \mapsto e^\alpha$ on real numbers,*

(ii) *the arithmetic operations $+, -, \times$, and $/$ on real numbers,*

(iii) *jumps conditioned on $>, \geq, <, \leq, =$, and $\neq$ comparisons of real numbers, and*

(iv) *output $0, 1$.*

*If the $t$ operations include no more than $p$ in which the exponential function is evaluated, then the VC-dimension of $\mathcal{H}$ is $\mathcal{O}(m^2 p^2 + mp(t + \log mp))$.*

**Theorem 12** *Let $\mathfrak{C}_\lambda$ be the family of $\lambda$-semi-spherical Gaussian mixtures with $k$ components. Let $\mathcal{X} \subset \mathbb{R}^d$ and $m = k(d+1)(d+2)/2 - 1$. Define $f_\theta : \mathcal{X} \to [0, \infty)$ as*

$$f_\theta(x) = -\ln\left(\sum_{i=1}^k \frac{w_i}{Z(\theta)\sqrt{|2\pi\Sigma_i|}} \exp\left(-\frac{1}{2}\left\|\Sigma_i^{-1/2}(x - \mu_i)\right\|_2^2\right)\right),$$

*where $Z(\theta) = \sum_i \frac{w_i}{\sqrt{|2\pi\Sigma_i|}}$. Let $\mathcal{F} = \{f_\theta(x) \mid \theta \in \mathfrak{C}_\lambda \subset \mathbb{R}^m\}$. Then, $\dim \mathcal{F} \in \mathcal{O}(d^4 k^4)$.*

**Proof** Let $\theta \in \mathfrak{C}_\lambda$ and $r \in \mathbb{R}$. Define $h : \mathbb{R}^{m+1} \times \mathbb{R}^d \to \{0, 1\}$ such that $h_{\theta,r}(x) = 1$ iff $f_\theta(x) \geq r$. Let the corresponding function class be defined as

$$\mathcal{H} = \{h_{\theta,r}(\cdot) \mid h_{\theta,r} : \mathcal{X} \to \{0, 1\} \mid \theta \in \mathfrak{C}_\lambda \subset \mathbb{R}^m, r \in \mathbb{R}\}.$$

For a fixed query $\theta$, all $\Sigma_i^{-1}, \forall i \in [k]$ are well defined as $\theta \in \mathfrak{C}_\lambda$. Furthermore, $\tilde{w}_i = w_i/(Z(\theta)\sqrt{|2\pi\Sigma_i|}) \in \mathbb{R}, \forall i \in [k]$ and $\sum_{i=1}^k \tilde{w}_i = 1$. As shown in Figure 5 the function $h_{\theta,r}(x)$ can be evaluated using $t = \mathcal{O}(m)$ arithmetic operations out of which exactly $k + 1$ are evaluations of the exponential function. Furthermore, it can be evaluated without using the natural logarithm by computing $e^{-r}$. By Lemma 10 and Theorem 11 we have

$$\dim \mathcal{F} = \dim_{\text{VC}} \mathcal{H} \in \mathcal{O}(m^2 k^2 + mk(t + \log mk)) \in \mathcal{O}(k^4 d^4).$$

∎

The lower-bound of $\Omega(kd^2)$ was established by Akama and Irie (2011). It is an open problem whether this gap can be closed further in the general setting.

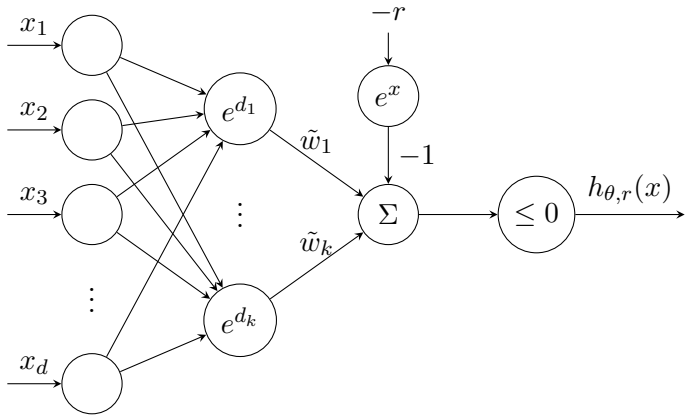

Figure 5: Feedforward network which calculates $h_{\theta,r}(x)$ whereby the immediate nodes compute the exponents of quadratic forms, and the resulting sum is compared to zero.

## A.3 Sufficient Coreset Size

Given the pseudo-dimension $\dim \mathcal{F}$ and some upper bound on the total sensitivity $\mathfrak{S}$ we may bound the coreset size by using the following theorem from Bachem et al. (2017b).

**Theorem 13 (Coreset size)** *Let $\varepsilon > 0$ and $\delta \in (0,1)$. Let $\mathcal{X}$ be a weighted data set, $\mathcal{Q}$ the set of all possible queries and $f_Q(x) : \mathcal{X} \times \mathcal{Q} \to \mathbb{R}_{\geq 0}$ a cost function. Let $s(x) : \mathcal{X} \to \mathbb{R}_{\geq 0}$ denote any upper bound on the sensitivity $\sigma(x)$ and define $S = \sum_{i=1}^{n} \mu_{\mathcal{X}}(x)s(x)$. Let $\mathfrak{C}$ be a sample of $m$ points from $\mathcal{X}$ with replacement where each point $x \in \mathcal{X}$ is sampled with probability $q(x) = \frac{\mu_{\mathcal{X}}(x)s(x)}{S}$ and each point $x \in \mathfrak{C}$ is assigned the weight $\mu_{\mathfrak{C}}(x) = \frac{\mu_{\mathcal{X}}(x)}{mq(x)}$. Let $\mathcal{F} = \left\{ \frac{\mu_{\mathcal{X}}(\cdot)f_Q(\cdot)}{\mathrm{cost}(\mathcal{X},Q)Sq(\cdot)} \mid Q \in \mathcal{Q} \right\}$ and $d' = \dim \mathcal{F}$. Then, the set $\mathfrak{C}$ is an $\varepsilon$-coreset of $\mathcal{X}$ with probability at least $1 - \delta$ for*

$$m \geq \frac{cS^2}{\varepsilon^2}\left(d' + \log\frac{1}{\delta}\right),$$

*where $c > 0$ is an absolute constant.*

A bound on the coreset size can also be obtained by applying the main theorem of Feldman and Langberg (2011), where one needs to bound the primal shattering dimension instead of the pseudo-dimension. For a detailed discussion of the effects introduced by this difference we refer the reader to Bachem et al. (2017a).

Now we are ready to present the proof of the Theorem 2 which states that, under natural assumptions, we can uniformly approximate the log-likelihood of the model trained on the coreset and the likelihood of the model trained on the full data set as $\varepsilon \to 0$.

**Theorem 2** *Let $\mathcal{X} \subset \mathbb{R}^d$, $\delta \in (0,1)$, $\varepsilon \in (0,1/2)$, $k \in \mathbb{N}$, and $\lambda \in (0,1)$. Let $\mathcal{B}_1, \ldots, \mathcal{B}_p$ be the outputs of $p = \lceil \log_2 1/\delta \rceil$ independent runs of Algorithm 2 with input $\mathcal{X}$ and $k$. Let $\mathcal{C}$ be the output of Algorithm 1 with $\alpha = 16(\log_2 k + 2)$, $\mathcal{B}^\star = \mathrm{argmin}_{i \in \{1,\ldots,p\}} \phi(\mathcal{X}, \mathcal{B}_i)$ and coreset size*

$$m \geq c \cdot \frac{d^4 k^6 + k^2 \log\frac{1}{\delta}}{\lambda^4 \varepsilon^2},$$

*where $c > 0$ is an absolute constant. Then, with probability at least $1 - \delta$, the set $\mathcal{C}$ is a $(k, \varepsilon)$-coreset of $\mathcal{X}$.*

**Proof** Since all $p$ runs are independent, it holds that

$$\mathbb{P}\left[\phi(\mathcal{X}, \mathcal{B}^\star) > t\right] = \mathbb{P}\left[\min_{i \in \{1, \ldots, p\}} \phi(\mathcal{X}, \mathcal{B}_i) > t\right] = \Pi_{i=1}^p \mathbb{P}\left[\phi(\mathcal{X}, \mathcal{B}_i) > t\right] \leq \left(\frac{\mathbb{E}[\phi(\mathcal{X}, \mathcal{B}_1)]}{t}\right)^p$$

by Markov's inequality. By Theorem 5 of Arthur and Vassilvitskii (2007), it holds that

$$\mathbb{E}[\phi(\mathcal{X}, \mathcal{B}_1)] \leq 8(\log_2 k + 2)\phi(\mathcal{X}, OPT).$$

Hence, for $p = \lceil \log_2 1/\delta \rceil$, with probability at least $1 - \delta$,

$$\phi(\mathcal{X}, \mathcal{B}^\star) \leq 16(\log_2 k + 2)\phi(\mathcal{X}, OPT).$$

By Lemma 7 we have that

$$s(x) = |\mathcal{X}|\frac{2}{\lambda^2}\left(\frac{\alpha \mathrm{d}(x, \mathcal{B})^2}{\sum_{x' \in \mathcal{X}} \mathrm{d}(x', \mathcal{B})^2} + \frac{2\alpha}{|\mathcal{X}_j|}\frac{\sum_{x' \in \mathcal{X}_j} \mathrm{d}(x', \mathcal{B})^2}{\sum_{x' \in \mathcal{X}} \mathrm{d}(x', \mathcal{B})^2} + \frac{2}{|\mathcal{X}_j|}\right) \geq \sigma(x)$$

for each $x \in \mathcal{X}$ and

$$\mathfrak{S} \leq \frac{1}{|\mathcal{X}|}\sum_{x \in \mathcal{X}} s(x) = \frac{4\alpha + 2\beta}{\lambda^2} \in \mathcal{O}\left(\frac{k}{\lambda^2}\right) \tag{5}$$

since $\alpha = \mathcal{O}(\log_2 k)$ and $\beta = k$. We conclude the proof by instantiating Theorem 13 with $\mu_{\mathcal{X}}(x) = 1/|\mathcal{X}|$, Theorem 12 and the total sensitivity bound (5). ∎

## A.4 Directly Approximating the Log-Likelihood

Under additional assumptions on the eigenvalues we can derive a stronger result directly relating the approximated log-likelihood with the true likelihood.

**Theorem 14** *Let the conditions of Theorem 2 hold. If $\prod_{\lambda_j \in \mathrm{spec}(\Sigma_i)} \lambda_j \geq \frac{1}{(2\pi)^d}$ we have*

$$|\mathcal{L}(\mathcal{X} \mid \theta) - \mathcal{L}(\mathcal{C} \mid \theta)| \leq \varepsilon \mathcal{L}(\mathcal{X} \mid \theta).$$

**Proof** By Theorem 2, for $\alpha = 16(\log_2 k + 2)$, $\beta = k$, $m \in \Theta\left(d^4 k^6 \lambda^{-4} \varepsilon^{-2}\right)$, with probability at least $1 - \delta$, it holds that

$$(1 - \varepsilon)\operatorname{cost}(\mathcal{X}, \theta) \leq \operatorname{cost}(\mathcal{C}, \theta) \leq (1 + \varepsilon)\operatorname{cost}(\mathcal{X}, \theta).$$

By assumption that all eigenvalues are sufficiently large, namely $\prod_{\lambda_j \in \mathrm{spec}(\Sigma_i)} \lambda_j \geq \frac{1}{(2\pi)^d}$, for all components $i$, the log-normalizer $\ln Z(\theta)$ is negative since

$$Z(\theta) = \sum_i \frac{w_i}{\sqrt{|2\pi\Sigma_i|}} \leq \max_i \frac{1}{\sqrt{|2\pi\Sigma_i|}} = \max_i \frac{1}{\sqrt{(2\pi)^d \prod_{\lambda_j \in \mathrm{spec}(\Sigma_i)} \lambda_j}} \leq 1.$$

Hence,

$$\mathcal{L}(\mathcal{C} \mid \theta) = -n \ln Z(\theta) + \text{cost}(\mathcal{C}, \theta) \leq -n \ln Z(\theta) + (1 + \varepsilon) \, \text{cost}(\mathcal{X}, \theta)$$
$$\leq (1 + \varepsilon)\big(-n \ln Z(\theta) + \text{cost}(\mathcal{X}, \theta)\big) = (1 + \varepsilon)\mathcal{L}(\mathcal{X} \mid \theta),$$

and similarly,

$$\mathcal{L}(\mathcal{C} \mid \theta) = -n \ln Z(\theta) + \text{cost}(\mathcal{C}, \theta) \geq -n \ln Z(\theta) + (1 - \varepsilon) \, \text{cost}(\mathcal{X}, \theta)$$
$$\geq (1 - \varepsilon)\big(-n \ln Z(\theta) + \text{cost}(\mathcal{X}, \theta)\big) = (1 - \varepsilon)\mathcal{L}(\mathcal{X} \mid \theta).$$

∎

## Appendix B. Convergence of Weighted EM for Gaussian Mixtures

We present the EM update equations for fitting a weighted set of points using Algorithm 4. Since we are interested in an MLE we begin by stating the necessary conditions for a stationary point of

$$\mathcal{L}(\mathcal{C} \mid \theta) = -n \ln Z(\theta) + \text{cost}(\mathcal{C}, \theta) = -\sum_i \gamma_i \ln P(\mathbf{x}_i' \mid \theta).$$

By assumption, all covariance matrices are non-singular. Taking the derivative of $\mathcal{L}(\mathcal{C} \mid \theta)$ with respect to $\mu_i$ and $\Sigma_i$ and setting it equal to zero yields

$$\mu_i = \frac{1}{N_i} \sum_{j=1}^{n} \eta_{i,j} x_j \quad \text{and} \quad \Sigma_i = \frac{1}{N_i} \sum_{j=1}^{n} \eta_{i,j} (x_j - \mu_i)(x_j - \mu_i)^T$$

where

$$N_i = \sum_{j=1}^{n} \eta_{i,j} \gamma_j \quad \text{and} \quad \eta_{i,j} = \gamma_j \frac{w_i \mathcal{N}(\mathbf{x}_j'; \mu_i, \Sigma_i)}{\sum_\ell w_\ell \mathcal{N}(\mathbf{x}_j'; \mu_\ell, \Sigma_\ell)}.$$

To find the mixing weights we minimize the negative log-likelihood under the constraint

$$\sum_{i=1}^{k} w_i = 1, w_i \geq 0, i = 1, \dots, k.$$

To this end we introduce a Lagrange multiplier $\lambda$ and minimize $\mathcal{L}(\mathcal{C} \mid \theta) + \lambda(\sum_{i=1}^{k} w_i - 1)$. Setting the derivative with respect to $w_i$ to zero yields

$$w_i = \frac{N_i}{\sum_{j=1}^{n} \eta_{i,j}}$$

As expected, the only difference to the non-weighted version of the EM algorithm is that the updates are now scaled proportionally to the weight of each point. As shown in Dempster et al. (1977) this algorithm will converge to a stationary point.